\pdfoutput=1
\documentclass[11pt]{article}

\usepackage[preprint]{coling}
\usepackage[T1]{fontenc}

\usepackage{times}
\usepackage{latexsym}

\usepackage[utf8]{inputenc}

\usepackage{graphicx}

\usepackage{tabularx}
\usepackage{abstract}

\usepackage{microtype}

\usepackage{inconsolata}

\title{From Arabic Text to Puzzles: LLM-Driven Development of Arabic Educational Crosswords}

\author{
\textbf{Kamyar Zeinalipour\textsuperscript{1}},
\textbf{Mohamed Zaky Saad\textsuperscript{1}},
\textbf{Marco Maggini\textsuperscript{1}},
\textbf{Marco Gori\textsuperscript{1}},
\\
\\
  \textsuperscript{1}University of Siena, DIISM, Via Roma 56, 53100 Siena, Italy
\\
  \small{
   \textbf{Correspondence:} \href{kamyar.zeinalipour2@unisi.it}{kamyar.zeinalipour2@unisi.it}
  }
}  

\begin{document}
\maketitle
\begin{abstract}
We present an Arabic crossword puzzle generator from a given text that utilizes advanced language models such as \texttt{GPT-4-Turbo}, \texttt{GPT-3.5-Turbo } and  \texttt{Llama3-8B-Instruct}, specifically developed for educational purposes, this innovative generator leverages a meticulously compiled dataset named \textit{Arabic-Clue-Instruct} with over 50,000 entries encompassing text, answers, clues, and categories. This dataset is intricately designed to aid in the generation of pertinent clues linked to specific texts and keywords within defined categories.
This project addresses the scarcity of advanced educational tools tailored for the Arabic language, promoting enhanced language learning and cognitive development. By providing a culturally and linguistically relevant tool, our objective is to make learning more engaging and effective through gamification and interactivity. Integrating state-of-the-art artificial intelligence with contemporary learning methodologies, this tool can generate crossword puzzles from any given educational text, thereby facilitating an interactive and enjoyable learning experience. This tool not only advances educational paradigms but also sets a new standard in interactive and cognitive learning technologies. The model and dataset are publicly available.
\footnote{\url{https://github.com/KamyarZeinalipour/Arabic-Text-to-Crosswords}}
\footnote{\url{https://huggingface.co/Kamyar-zeinalipour/Llama3-8B-Ar-Text-to-Cross}}
\footnote{\url{https://huggingface.co/datasets/Kamyar-zeinalipour/Arabic-Clue-Instruct}}
\end{abstract}

% no keywords

% For peer review papers, you can put extra information on the cover
% page as needed:
% \ifCLASSOPTIONpeerreview
% \begin{center} \bfseries EDICS Category: 3-BBND \end{center}
% \fi
%
% For peerreview papers, this IEEEtran command inserts a page break and
% creates the second title. It will be ignored for other modes.

\section{Introduction} \label{sec:intoduction}
Crossword puzzles, traditionally enjoyed for their challenge and entertainment value, are increasingly recognized for their educational potential. These puzzles facilitate learning in multiple disciplines, such as history, science, and linguistics, and are particularly effective in enhancing vocabulary and spelling skills \cite{orawiwatnakul2013crossword,bella2023improving,dzulfikri2016application}. Their capacity to engage and educate simultaneously makes them invaluable in pedagogical contexts.\\
In language acquisition and the mastery of specialized terminology, crossword puzzles stand out as exceptional tools. They offer an interactive learning experience that promotes both the retention of technical jargon and general language skills \cite{nickerson1977crossword,yuriev2016crossword,sandiuc2020use}. This dynamic learning approach supports the cognitive development of learners by improving critical thinking abilities and memory retention \cite{kaynak2023effect,dzulfikri2016application,mueller2018testing,zirawaga2017gaming,bella2023improving,zamani2021use,dol2017gpbl}.\\
The integration of advanced technologies, such as Natural Language Processing (NLP), further enhances the effectiveness of educational crossword puzzles. The advent of Large Language Models (LLMs) has particularly revolutionized the creation of Arabic educational crosswords, providing sophisticated, context-appropriate clues that significantly enrich the learning experience.\\
This paper introduces a novel application that leverages LLMs to produce tailored educational crossword puzzles from given text in a chosen category for Arabic learners. The application creates high-quality clues and answers, integrating user-provided texts or keywords through fine-tuning techniques. which can be utilized by educators to develop more engaging and effective instructional methodologies for learners.\\
Moreover, to support the development and assessment of this tool, a new dataset has been compiled named \textit{Arabic-Clue-Instruct}, which will be disseminated to the scientific community. This dataset, accessible in an open-source format, consists of categorized texts in Arabic, each corresponding with clues and keywords in various learning categories. This initiative aims to streamline the creation of educational crosswords and facilitate their adoption in educational settings.\\
The structure of the paper is as follows. Section \ref{sec:relatedworks} provides a detailed review of the relevant literature. Section \ref{sec:Methodology} contains the data set collection methodology and the curation process. The computational methodologies used are also elucidated. Section \ref{sec:Experiments} discusses the results of our experimental evaluations. Finally, Section \ref{sec:conclusions} offers concluding remarks on the results and implications of our study. and Section \ref{sec:limitation} outlines the limitations of this study. 
\section{Related Works} \label{sec:relatedworks}
The field of crossword puzzle generation has attracted significant academic interest due to its complex nature. Researchers have experimented with a wide array of approaches including the use of vast lexical databases and embracing the potential of modern Large Language Models (LLMs) \cite{harnessingllms, automatingturkish}, which accommodate innovative methods like fine-tuning and zero/few-shot learning.\\
Pioneer works in this domain have been diverse and innovative. For example, Rigutini and his colleagues. successfully employed advanced natural language processing (NLP) techniques to automatically generate crossword puzzles directly from sources based on the Internet, marking a fundamental advancement in this field~\cite{rigutini2008fully, rigutini2012automatic, turkishgenerator}. In addition, Ranaivo and his colleagues. devised an approach based on NLP that utilizes text analytics alongside graph theory to refine and pick clues from texts that are topic-specific~\cite{ranaivo2013automatic}. Meanwhile, aimed at catering to Spanish speakers, Esteche and his group developed puzzles utilizing electronic dictionaries and news articles for crafting clues~\cite{esteche2017automatic}. On another front, Arora and his colleagues. introduced SEEKH, which merges statistical and linguistic analyzes to create crossword puzzles in various Indian languages, focusing on keyword identification to structure the puzzles~\cite{arora2019automatic}.\\
Recent advances have included the efforts of Zeinalipour and his colleagues, who demonstrated how large-scale language models can be used to develop crossword puzzles in lesser-supported languages, including English, Arabic, Italian and Turkish showcasing the broad applicability of computational linguistics in generating puzzles that are both engaging and linguistically rich~\cite{zeinalipour2023arabicros, zeinalipour2023italian, zeinalipour2023building,turkishgenerator,zeinalipour2024harnessing}.
In \cite{zugarini2024clue} they suggest a method for creating educational crossword clues and text datasets in English.
In the Arabic version of generating crossword puzzles from text \cite{zeinalipour2023arabicros}, a few-shot learning approach was utilized, employing large language models (LLMs) as they are. However, in our current project, we have introduced a dataset specifically designed for this task in Arabic. Additionally, we have developed open-source models fine-tuned to better accommodate and enhance performance for this specific application.\\
However, the unique characteristics and challenges presented by the Arabic language have remained largely unexplored in the context of the generation of crossword puzzles from a given text. This study breaks new ground by explicitly employing cutting-edge language modeling techniques to create Arabic crossword puzzles from a given text. This approach not only fills a gap within the scholarly literature, but also enhances the tools available for language-based educational initiatives, thus advancing the field of Arabic crossword puzzle generation.
\section{Methodology}\label{sec:Methodology}
\begin{figure*}[ht!]
    \centering
       \includegraphics[width=1\textwidth]{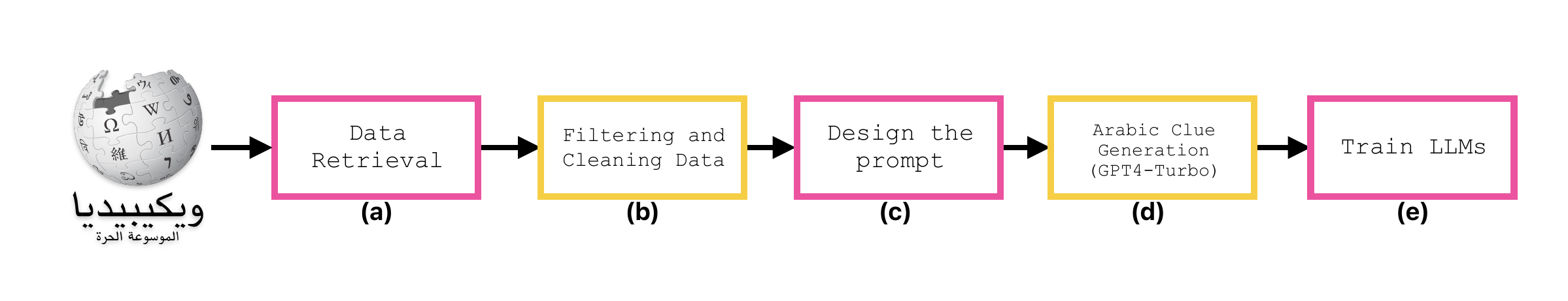}
    \caption{The methodology employed in this study is illustrated in this figure and includes the following steps: (a) Gathering data from Arabic Wikipedia. (b) Refining and filtering the data to enhance quality by eliminating content that is either too brief or excessively detailed. (c) Developing prompts for creating educational Arabic crossword clues derived from the educational content. (d) Employing \texttt{GPT4-Turbo} to generate Arabic crossword clues using the refined data and specifically crafted prompts. (e) Fine-tuning Large Language Models (LLMs) to more effectively produce Arabic clues tailored to the given context.}
    \label{fig:fig1}
\end{figure*}

This paper outlines the development of an innovative tool: an automated system to generate educational crossword puzzles from given text in Arabic, driven by Large Language Models (LLM). We introduce the \textit{Arabic-Clue-Instruct} dataset, which comprises a curated selection of Arabic texts across several educational fields, including Chemistry, Biology, Geography, Philosophy, and History, among others.Our research is not confined to the specific category utilized in this study; it can be generalized to encompass a broader range of categories. This generalizability is attributed to the inherent capabilities of LLMs.This dataset served as the foundation for creating tailored crossword clues for each subject area.\\
An exhaustive iterative procedure enabled us to enrich the dataset, ensuring a diversified representation of Arabic educational material and specific crossword clues to correspond with each thematic area.\\
Our primary objective was to engineer Arabic crossword puzzles directly from textual content, employing the \textit{Arabic-Clue-Instruct} dataset. We accomplished this by initially extracting the main context and the suitable keywords then generating human-evaluated clues using GPT-4-Turbo \ref{sec:datprompt} then fine-tuning a suite of LLMs to adeptly generate accurate and pertinent crossword clues. The models employed in this undertaking include \texttt{Llama3-8B-Instruct} and \texttt{GPT3.5-Turbo }.\\
The subsequent sections of this paper will explore the detailed techniques utilized in the creation of this dataset and the customization of LLMs. This development is expected to enhance learning experiences and outcomes in Arabic education. The comprehensive methods employed in this project are visualized in Figure \ref{fig:fig1}.

\subsection{\textit{Arabic-Clue-Instruct}} 
\label{sec:dataset}

\paragraph{Data Collection Methodology} \label{sec:data_acquisition}
We initiated our data acquisition by extracting the initial sections of Arabic Wikipedia articles, specifically focusing on the prominently bolded keywords included in the Introduction section that correspond to the article's primary focus and other significant terms. In addition to this keyword-focused examination, we acquire essential metadata for each article, including metrics such as view counts, relevance assessments, condensed narrative overviews, key headlines, associative terms, categorization, and URLs.\footnote{\url{https://en.wikipedia.org/wiki/Wikipedia:Lists_of_popular_pages_by_WikiProject} Wikipedia: Lists of popular pages by WikiProject}
 This procedure benefits from the consistent structural format of Arabic Wikipedia, especially exploiting the information-dense introductory segments to methodically outline and extract the core ideas needed for a comprehensive data repository.

\paragraph{Data Refinement Techniques} \label{sec:data_filtering}
To enhance data quality through precision filtering, our strategy involves several critical steps. Initially, article selection is guided by measurements of popularity and relevance. We then discard articles that are either excessively long or significantly short in content by excluding those with a word count of less than 50 to ensure there is more room for clue generation. Furthermore, any associations with keywords that are more than two words long are removed from consideration to enhance the quality of the generated crossword puzzle. In the final step of our filtering process, we exclude keywords that either fall outside of the 3 to 20-character limit or include special characters and numerals. These filtering techniques are fundamental in maintaining the integrity and applicability of our dataset.

\begin{figure}[ht!]
    \centering
    \includegraphics[width=\columnwidth]{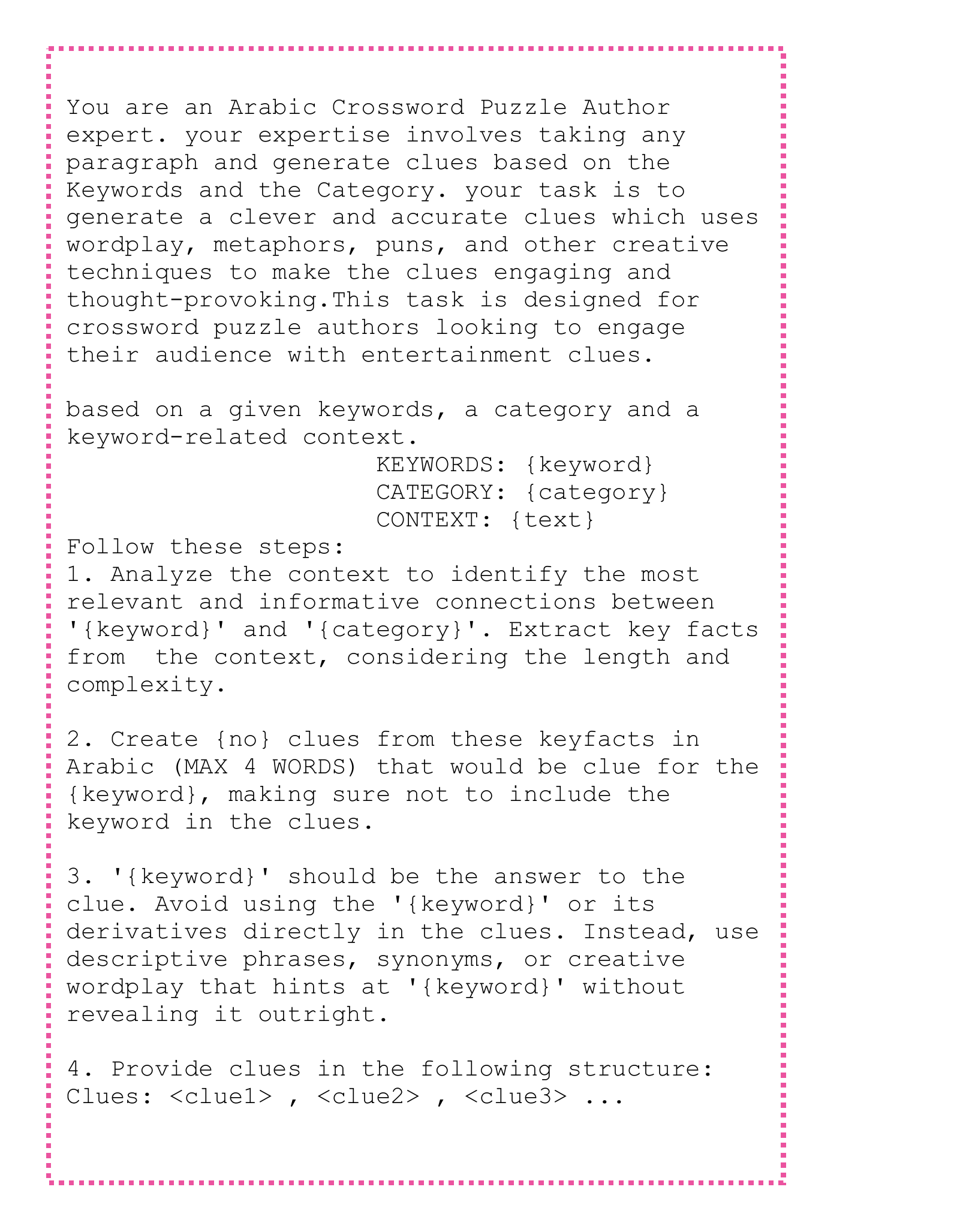}
    \caption{Prompt used in the study.}
    \label{fig:prompt}
\end{figure}

\paragraph{Development of the Prompting Mechanism.} \label{sec:datprompt}
Creating a specific prompt was crucial for generating Arabic crossword clues from the given text using GPT-4-Turbo. These clues, which will be part of the main dataset for fine-tuning, relied heavily on Wikipedia articles to maintain topical relevance.
 This prompt was meticulously designed to facilitate the formulation of clues that were not only informative but also engaging, by weaving essential details and contextual background derived from the articles into the clues. The prompt was designed using the latest prompt engineering best practices and after trying different structures and prompting in Arabic and in English. The prompt utilized in this study is illustrated in Figure \ref{fig:prompt}.

\paragraph{Formulation of Educational Arabic Clues.} \label{sec:datagen}
Influenced by the \textsc{self-instruct} framework~\cite{wang2022self}, our approach leverages Large Language Models to automate the creation of educational crossword clues in Arabic from the given text. Our strategy is distinctive in its thorough integration of contextual information with the generated clues. For this purpose, we utilized \texttt{GPT-4-Turbo} an advanced version of LLMs known for its superior efficiency and capabilities. The culmination of our methodology involves employing carefully curated content, keywords, categories, and prompts as the foundational elements for crafting custom Arabic educational clues that meet our specific requirements.

\paragraph{Overview of the \textit{Arabic-Clue-Instruct} Dataset}
% Incorporation of data filtration steps
% Utilization of GPT to ensure accurate category assignment

We initiated our study by downloading approximately 211,000 articles from Arabic Wikipedia. This number was reduced to 14,497 suitable entries following a stringent content filtration process. These pages span 20 distinct thematic categories and form the basis of our curated dataset which features a mix of textual content and associated keywords.The data set was divided into 14,000 for training and 497 for testing. \\
To enrich the dataset further, we utilized the capabilities of \texttt{GPT-4-Turbo}, generating a diverse set of at least three clues per Wikipedia entry based on the length of the text. This effort resulted in a compilation of 54,196 unique clues, as detailed in Table \ref{tab:stat2}.
\begin{table*}[ht]
    \centering
    \begin{tabular}{ccc}
        \hline
        \multicolumn{3}{c}{\textbf{\textit{Arabic-Clue-Instruct}}} \\
        \hline
        Total Content-Keyword Pairs & Number of Distinct Categories & Total Generated Clues\\
        \hline
        14,497  & 20 & 54,196  \\
        \hline
    \end{tabular}
    \caption{Overview of the \textit{Arabic-Clue-Instruct} Dataset}
    \label{tab:stat2}
\end{table*}

\begin{figure}
    \centering
       \includegraphics[width=\columnwidth]{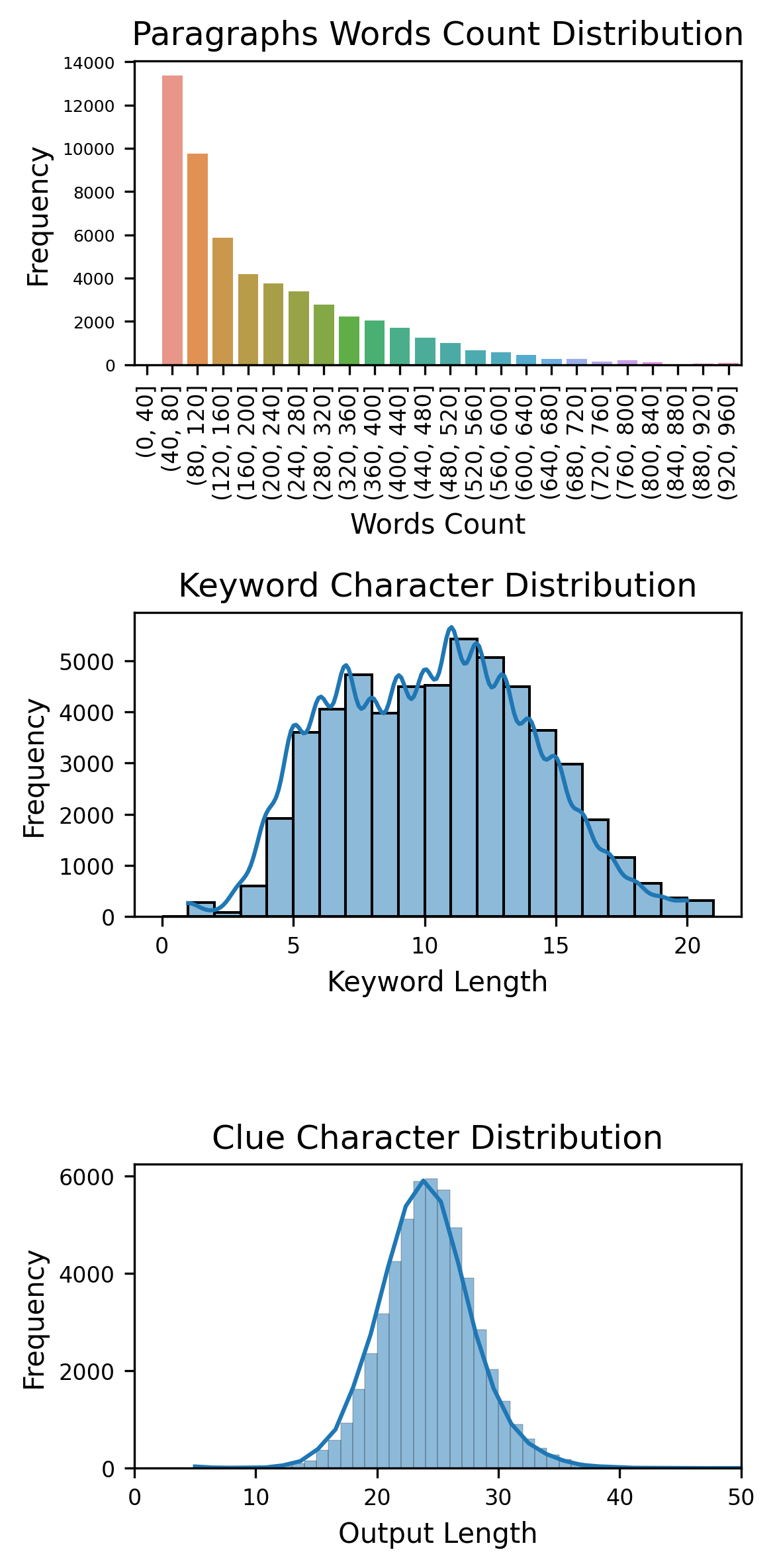}
    \caption{Word and Character Length Distributions for Contexts, Outputs, and Keywords.}
    \label{fig:dataset_distrubtions}
\end{figure}

\begin{figure}
    \centering 
   \includegraphics[width=\columnwidth]{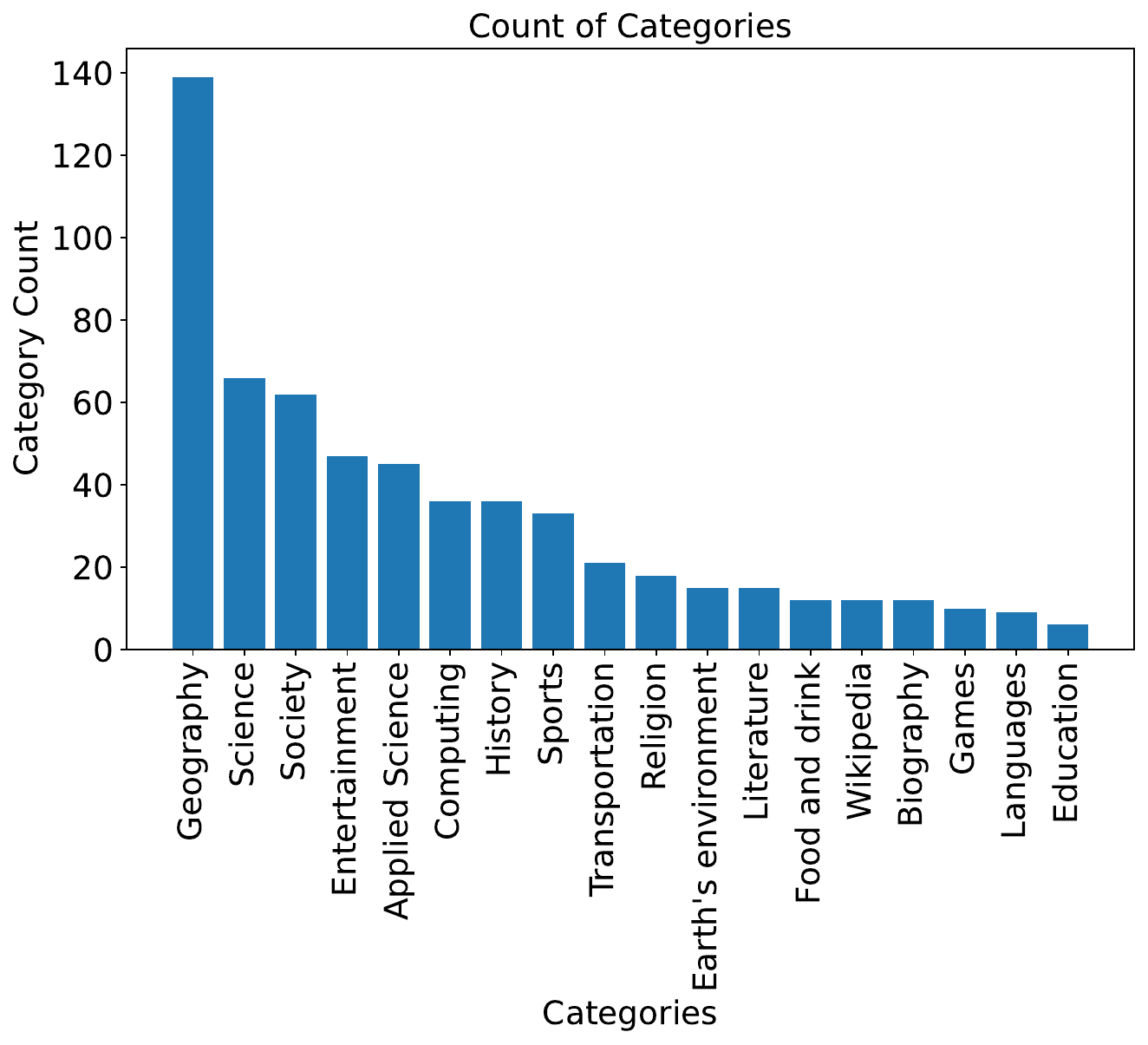}
    \caption{Bar Plot Showing the Frequency of Twenty Categories within the Dataset.}
    \label{fig:topics-distr}
\end{figure}

A deeper analysis of the dataset reveals variability in context length, ranging from 75 to 1000 words, with most texts falling between 40 and 200 words. The clue-generation process typically yields clues between 20 and 30 characters in length. The character count for keywords is restricted to between 2 and 20 characters, and the majority of them are between 5 to 15 characters long. The distribution of text word counts, keyword occurrences, and clue character lengths is depicted in Figure~\ref{fig:dataset_distrubtions}.\\
Figure \ref{fig:topics-distr} shows the distribution of data across various categories. Notably, the categories of "Geography", "Science", and "Society" dominate the dataset, whereas topics such as "Games", "Languages", and "Education" are comparatively underrepresented.


\paragraph{Evaluating quality of the \textit{Arabic-Clue-Instruct} Dataset}
\label{framework}
Producing accurate and engaging Arabic educational crossword clues is inhibited by the absence of a reference corpus, making it difficult to draw comparisons using standard measures, such as ROUGE scores. As typical metrics for comparison are not available, our evaluation strategy needed to adapt uniquely to the task requirements. Specifically, effective clues should represent contextually accurate paraphrases of text information. To accommodate this, we adopted an extractive method, using the ROUGE-L score to gauge the adequacy of clues in reflecting the input context. By comparing input sentences to the generated clues, the evaluation aimed to attain high scores to ensure strict adherence to the original text, minimizing irrelevant content and avoiding clues that merely replicate the input or improperly introduce the target keyword. Results indicated a substantial connection between the context and the clues, with an average ROUGE-L score of 0.0278, detailed results are provided in Table \ref{tab:rouge_score_gpt4}.

\begin{table*}[ht]
    \centering
\begin{tabular}{c c c c c} 
 \hline
 \textbf{Candidates} & \textbf{ROUGE-1} & \textbf{ROUGE-2} & \textbf{ROUGE-L} \\
\hline
        Text vs GPT4-Turbo clues & 0.0281 & 0.0055 & 0.0278 \\
 \hline
\end{tabular}
    \caption{Mean ROUGE Scores for GPT4 vs Text}
     \label{tab:rouge_score_gpt4}
\end{table*}
Considering that the ROUGE score merely compares the similarity between the n-grams of the generated clues and the reference text, it is not a reliable metric and does not provide any assessment of the semantic quality of the generated clues. However, it gives some thoughts about the generated clues.\\
In addition, the integrity of the generated clues was further examined through human evaluations conducted by experts in the Arabic language. A randomly selected subset of clues was assessed, consisting of 200 articles, each containing a maximum of three clues. This evaluation used a five-level criteria system, analogous to the methodology used by \cite{wang2022self}, which detailed the parameters as follows:
\begin{itemize}
    \item \texttt{RATING-A}: The clue is coherent and valid, aligning correctly with the given context, answer, and specified category.
    \item \texttt{RATING-B}: This clue, while generally acceptable, exhibits slight discrepancies mainly characterized by a tenuous link to the category.
    \item \texttt{RATING-C}: Here, the clue relates directly to the answer but retains either a vague connection to the context or seems too broad.
    \item \texttt{RATING-D}: The clue fails by being irrelevant or incorrect in relation to the answer or the context.
    \item \texttt{RATING-E}: The clue is deemed unacceptable because it directly contains the answer or a variation of it.
\end{itemize}

The evaluation was made by a Native Arabic speaker who followed the criteria of evaluating based on the criteria mentioned above, As "Rating-A" was given to the good clues without any mistakes, "Rating-B" was given to the clues with minor mistakes but the clue is generally accepted, 'Rating-C' was given to the clue which has mistakes or the connection isn't direct to the article but can still be understandable,  "Rating-D" was given to clues that were entirely incorrect due to nonsensical words or irrelevance to the answer, and "Rating-E" to clues that contained the answer itself, making them unacceptable regardless of other qualities. 
The data and code used for evaluating the human annotation user interface (UI) are accessible on GitHub.\footnote{\url{https://github.com/KamyarZeinalipour/HumanAnnotation-UI-Ar-Text-To-Cross}}

\begin{figure}
    \centering 
   \includegraphics[width=\columnwidth]{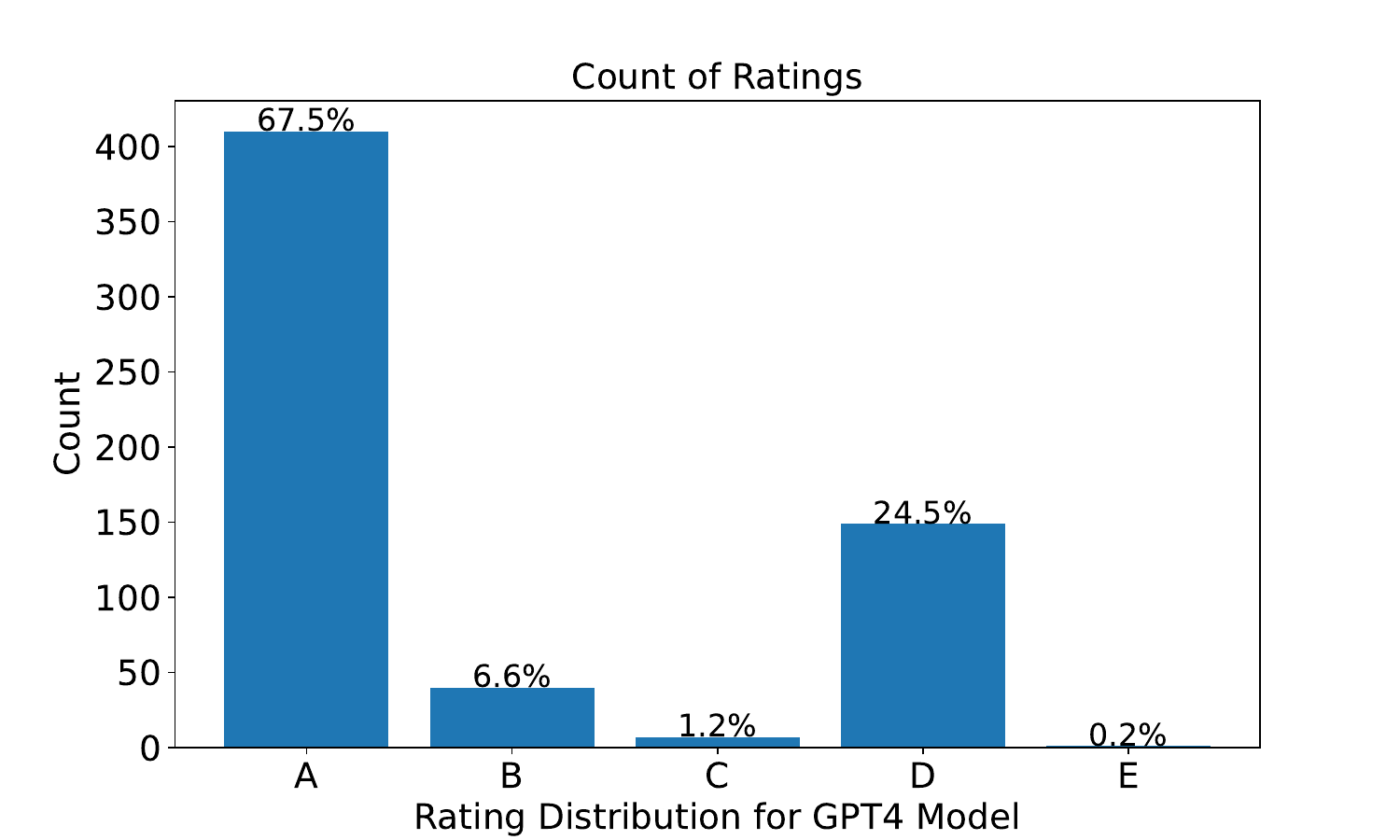}
    \caption{Bar Plot Showing the Frequency of GPT4 Ratings}
    \label{fig:count_of_ratings}
\end{figure}

The distribution of the evaluation outcomes is depicted in Figure \ref{fig:count_of_ratings} which also contains the percentage for each rating. These illustrate that the majority of the generated clues were of high quality with over 67.5\% rated as 'A' and only a small fraction rated as 'C', 'D', or 'E'.\\
 A qualitative analysis is available in Appendix \ref{appendix_2}. Appendix\ref{sec:appendix_a} provides examples of different ratings for various clues generated by the \texttt{GPT-4-Turbo}.\\
By utilizing both quantitative metrics and qualitative assessments, the study aimed to robustly validate the educational utility and contextual accuracy of the clues created for Arabic educational crosswords.

\subsection{Enhancing Arabic Educational Crossword Puzzle Generation Using LLMs}.

The study focus was the generation of crossword puzzle clues derived from Arabic texts, utilizing the advanced capabilities of Large Language Models (LLMs). The selected models for this task included \texttt{GPT-4-turbo}, \texttt{Llama3-8B-Instruct}, and \texttt{GPT-3.5-Turbo }, all of which are noted for their text generation skills and Arabic language support \cite{brown2020language,touvron2023llama}. These models excel in decoding and formulating complex language structures, making them ideal for our objectives.\\
The process started with a crucial stage of model fine-tuning using the \textit{Arabic-Clue-Instruct} dataset, rich in relevant content for our needs. This fine-tuning was critical to better equip the models to generate Arabic clues and to refine their ability to mirror the linguistic intricacies specific to the Arabic language used in educational contexts.\\
We applied meticulous parameter optimization to the fine-tuning phase to decrease task-specific errors. Our strategy had a two-fold purpose: to bolster the model’s understanding of the Arabic educational material and to ensure an accurate representation of the language in the clues. The Arabic language presents unique challenges due to its complex grammar and vocabulary, making this an intensive task.\\
By tailoring these sophisticated LLMs with a dedicated dataset, we tried to enhance their capability to produce crossword clues from given Arabic text that is not just linguistically coherent but also contextually relevant to educational settings.
\section{Experimental Results}\label{sec:Experiments}
In this section, we will detail the comprehensive experiments conducted in this study. Initially, we begin by describing the training setup utilized for training the LLMs using the \textit{Arabic-Clue-Instruct}. This includes training parameters and the computational resources employed.  Subsequently, we present the performance evaluations of the models, highlighting their effectiveness through automated metrics, with a particular focus on the ROUGE score. We will break down the results to show how the various configurations of the models compare and identify key areas where performance improvements are observed.  Following this, we delve into an in-depth analysis of the human evaluations conducted to assess the models' performance. This analysis will include the methodology adopted for human evaluation, criteria for assessment, and qualitative feedback from human reviewers. We will provide insights into aspects such as relevance, coherence, and the overall quality of the generated content, which are often beyond the scope of automated metrics.  Lastly, to illustrate the practical application of our proposed system, we provide an example of a crossword puzzle generated using the trained models.  Through this detailed examination, we aim to present a holistic understanding of the experiments and their outcomes, demonstrating the robustness and versatility of the proposed system.
\subsection{Training Setup}

For the \texttt{GPT-3.5-Turbo}, the training regimen was meticulously designed, employing a batch size of 16 and a learning rate of 0.01 across three training epochs. Similarly, \texttt{Llama3-8B-Instruct} was fine-tuned using LORA \cite{hu2021lora} with parameters set to $r=32$ and $\alpha=64$ over the course of three training epochs, maintaining a total batch size of 128. The initial learning rate for \texttt{Llama3-8B-Instruct} was configured at $3\times 10^{-4}$. For the inference phase, model distribution sampling was employed to generate clues for \texttt{Llama3-8B-Instruct}, with a temperature parameter set to 0.1. Additionally, the parameters for top-$p$ and top-$k$ sampling were adjusted to 0.95 and 50, respectively. The entirety of the experimental setup was conducted on a server equipped with dual NVIDIA A6000 GPUs, utilizing DeepSpeed \cite{rasley2020deepspeed} and FlashAttention 2 \cite{dao2023flashattention} technologies.
\subsection{Evaluation Results with the Automatic Metrics}

\begin{table*}
    \centering
\begin{tabular}{c c c c c} 
 \hline
 \textbf{Model} & \textbf{Model name} & \textbf{ROUGE-1} & \textbf{ROUGE-2} & \textbf{ROUGE-L} \\
 \hline
  	Base LLMs & GPT-3.5 & 0.0152 & 0.0011 & 0.0148 \\

        &LlaMa3-8b& 0.0063 & 0.0009 & 0.0063 \\
  \hline
 
	Fine-tuned LLMs  & GPT-3.5& \textbf{0.0405} &  \textbf{0.0045} &  \textbf{0.0405} \\ 

          & LlaMa3-8b & 0.0354 & 0.0030 & 0.0354 \\
    \hline
\end{tabular}
    \caption{Mean ROUGE Scores for Various Comparisons with gpt4 clues}
     \label{tab:rouge_scores}
\end{table*}

We assessed the similarity between various sets of clues generated by different models, as presented in Table \ref{tab:rouge_scores}, and those generated by the \texttt{GPT-4-Turbo} model on a test set containing 200 educational contexts. This assessment was performed using ROUGE scores. The results reveal that the fine-tuned \texttt{Llama3-8b-Instruct} and \texttt{GPT-3.5-Turbo} model achieves a higher similarity to \texttt{GPT-4-Turbo}. In contrast,\texttt{Llama3-8B-Instruct} base model display significantly lower similarity exhibiting minimal overlap. These findings underscore the effectiveness of fine-tuning in aligning \texttt{GPT-3.5-Turbo} and \texttt{Llama3-8B-Instruct} using the \textit{Arabic-Clue-Instruct} data set was increase the capability of models for generating the clues from Arabic educational text.

\subsection{Evaluation Results with the human evaluator }

\begin{figure*}
    \centering 
   \includegraphics[width=\textwidth]{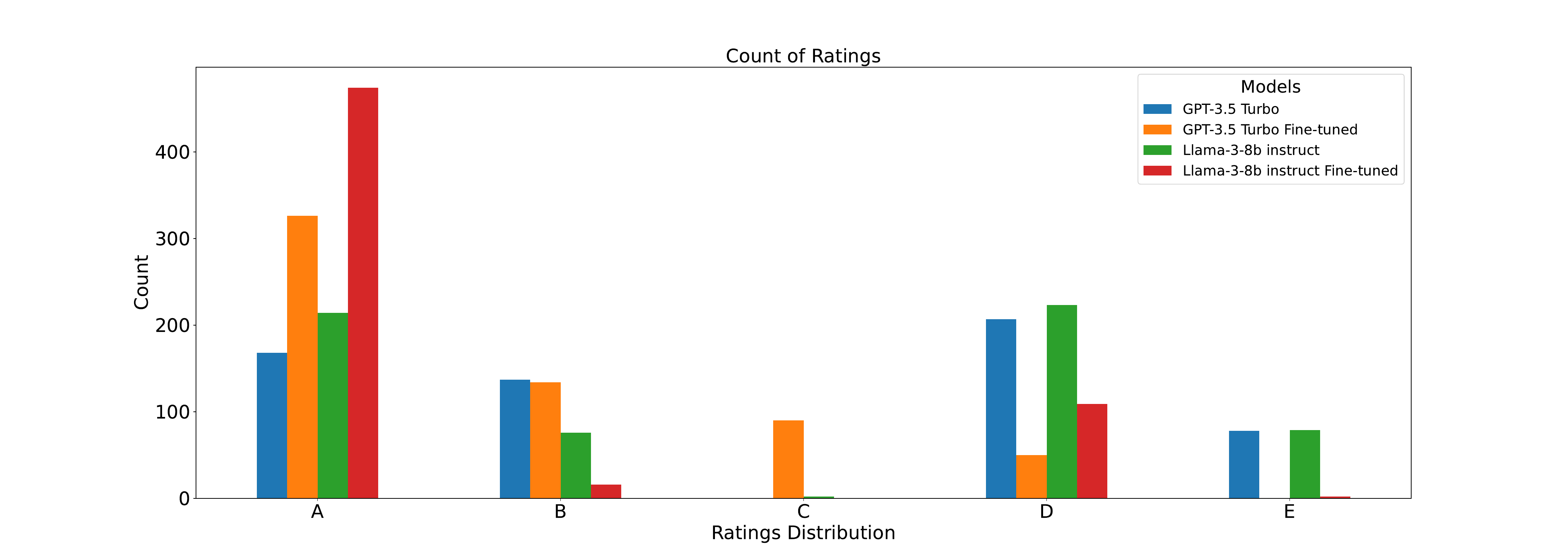}
    \caption{Bar Plot Displaying the Frequency of Ratings from Human Evaluation.}
    \label{fig:ratings-dist}
\end{figure*}

Since the ROUGE score is not a reliable metric for assessing semantic quality, we conducted a human evaluation based on the judgment framework detailed in a previous section \ref{framework} , A human evaluation was conducted on both the generated and base models using a data set of 200 Arabic contexts, each containing 3 clues. The resulting ratings are illustrated in Figure \ref{fig:ratings-dist} and summarized in Table \ref{tab:tab_perc}, which shows the percentage distribution of each model's ratings. We employed the same 5-level rating system detailed in Section \ref{sec:Methodology}.

 \begin{table*}
    \centering
    \begin{tabular}{ccccccc}
        \hline
     & \textbf{\texttt{GPT3.5-Turbo}} & \textbf{\texttt{GPT3.5-Turbo FT}}& \textbf{\texttt{Llama3-8B}} & \textbf{\texttt{Llama3-8B FT}}\\
    \hline
    \textbf{\# params}  &- & -& 8B&8B\\
    \hline
    &&\textbf{Ratings \%}&&&&\\
    \hline
    \textbf{A} & 28.47 & 54.33 & 36.02 & \textbf{78.86}  \\
    \textbf{B} &  23.22 & 22.33 & 12.79 & 2.66  \\
    \textbf{C} &  0 & 15.00 & 0.34 & 0  \\
    \textbf{D} & 35.08 &  8.33 & 37.54 & 18.13 \\
   \textbf{E} & 13.22 & 0 & 13.29 &  0.33 \\
    
    \hline
     \end{tabular}
    \caption{Assessing the percentage of human evaluation for the clues generated by LLMs.}
    \label{tab:tab_perc}

\end{table*}
The ratio of each rating can be seen in Table \ref{tab:tab_perc}. The provided table offers a comparative evaluation of language models' performance in generating Arabic clues from given text. Notably, both \texttt{GPT-3.5-Turbo} and \texttt{Llama3-8B-Instruct} models are assessed based on their base and fine-tuned configurations. Fine-tuning demonstrates a notable impact on model performance, as evidenced by the improvement of \texttt{GPT-3.5-Turbo FT} over its base counterpart. \texttt{Llama3-8B-Instruct} emerges as the top performer, achieving a remarkable 78.86\% in category "A" from Originally being 36.02\% in the Base Model which exceeds the \texttt{GPT-3.5-Turbo} in improving the performance after the fine tuning. The distribution of ratings showcases \texttt{Llama3-8B-Instruct} dominating category "A". These results underscore the efficacy of fine-tuning in enhancing model capabilities, particularly highlighted by \texttt{Llama3-8B-Instruct}'s performance with 8B parameters. Furthermore, after fine-tuning with the introduced dataset, the models' capability to generate Arabic clues from given text has significantly increased. This improvement also highlights the quality of the \textit{Arabic-Clue-Instruct} dataset. A qualitative analysis is available in Appendix \ref{appendix_2}. In Appendix \ref{sec:appendix_a}, you can find various examples of generated clue-answer pairs based on the provided text, along with their corresponding human ratings.

\begin{figure*}[ht!]
    \centering
       \includegraphics[width=0.9\textwidth]{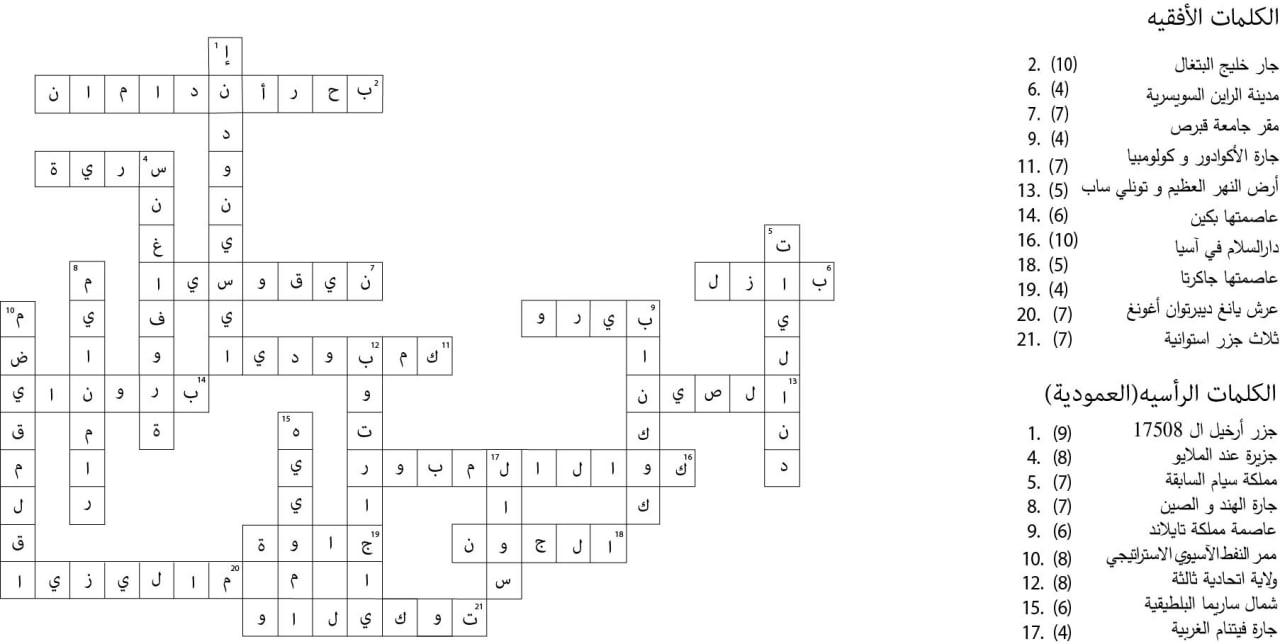}\\
    \caption{Crossword crafted using the proposed system.}
    \label{fig:crossword}
\end{figure*}

\paragraph{Generating Crossword Schema}
We investigated a methodology for generating Arabic crossword clues from categorized texts. This approach enables the creation of custom clues tailored to specific educational objectives. Following clue generation, educators can curate and select the most appropriate clues for constructing crosswords. An illustrative example of an Arabic crossword puzzle produced using this system is presented in Figure \ref{fig:crossword}.

\section{Conclusion}\label{sec:conclusions}

In this paper, we present a system for generating crossword clues from Arabic text. We created a dataset named \textit{Arabic-Clue-Instruct} that includes text, keywords, categories, and related crossword clues in Arabic, making it the first of its kind in this context. Utilizing this dataset, we fine-tuned two large language models (LLMs), namely \texttt{GPT-3.5-Turbo} and \texttt{Llama3-8B-Instruct}. Our results demonstrate a significant improvement in the models' capability to generate crossword clues from given text after fine-tuning. We have made the \textit{Arabic-Clue-Instruct} dataset, along with the fine-tuned models, publicly available. These tools can be especially useful for students and teachers to generate educational crossword puzzles from Arabic text, as they can implement this crossword puzzle generator as a supplementary learning tool across various subjects by integrating the puzzles directly into lesson plans or as homework assignments. For future work, we aim to develop models capable of generating various types of crossword clues, including fill-in-the-blank crossword clues, and to expand our approach by adding more diverse datasets and experimenting with methodologies for languages with limited resources.

\section{Limitations}\label{sec:limitation}

The Arabic crossword puzzle generator, while innovative, has several limitations. The \textit{Arabic-Clue-Instruct} dataset, despite its 50,000 entries, may not fully represent all Arabic dialects or recent language trends, potentially limiting its effectiveness across diverse Arabic-speaking regions.\\
Additionally, the tool's reliance on pre-defined categories might restrict its capacity to create puzzles for new or interdisciplinary topics, which could limit educators in customizing content for specific educational needs. Furthermore, varying levels of technological literacy among users might present challenges in using the tool effectively, potentially widening digital divides.\\
To address these limitations, future developments could focus on expanding and diversifying the dataset, enhancing the flexibility of puzzle generation, and ensuring the tool is accessible and user-friendly for a broad audience.
% Bibliography entries for the entire Anthology, followed by custom entries
%\bibliography{anthology,custom}
% Custom bibliography entries only
\bibliography{custom}

\appendix

\section{Example of clues and answers generated from text using different models.}
\label{sec:appendix_a}
\begin{figure*}
    \centering 
   \includegraphics[width=1\textwidth]{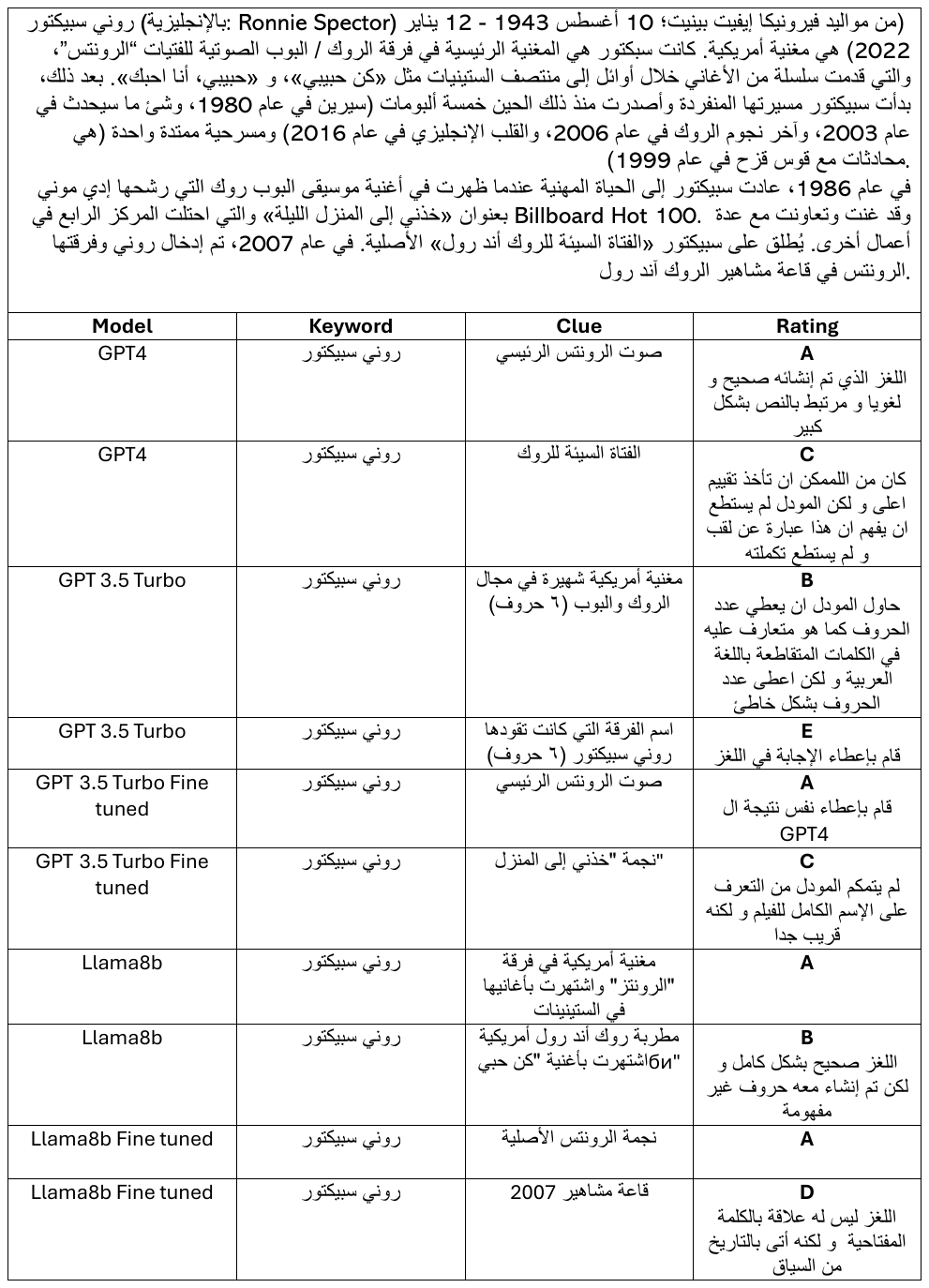}
    
    \caption{Example of Ronnie Spector article in Arabic}\label{fig:Example1_AR}
\end{figure*}

\begin{figure*}
    \centering 
   \includegraphics[width=1\textwidth]{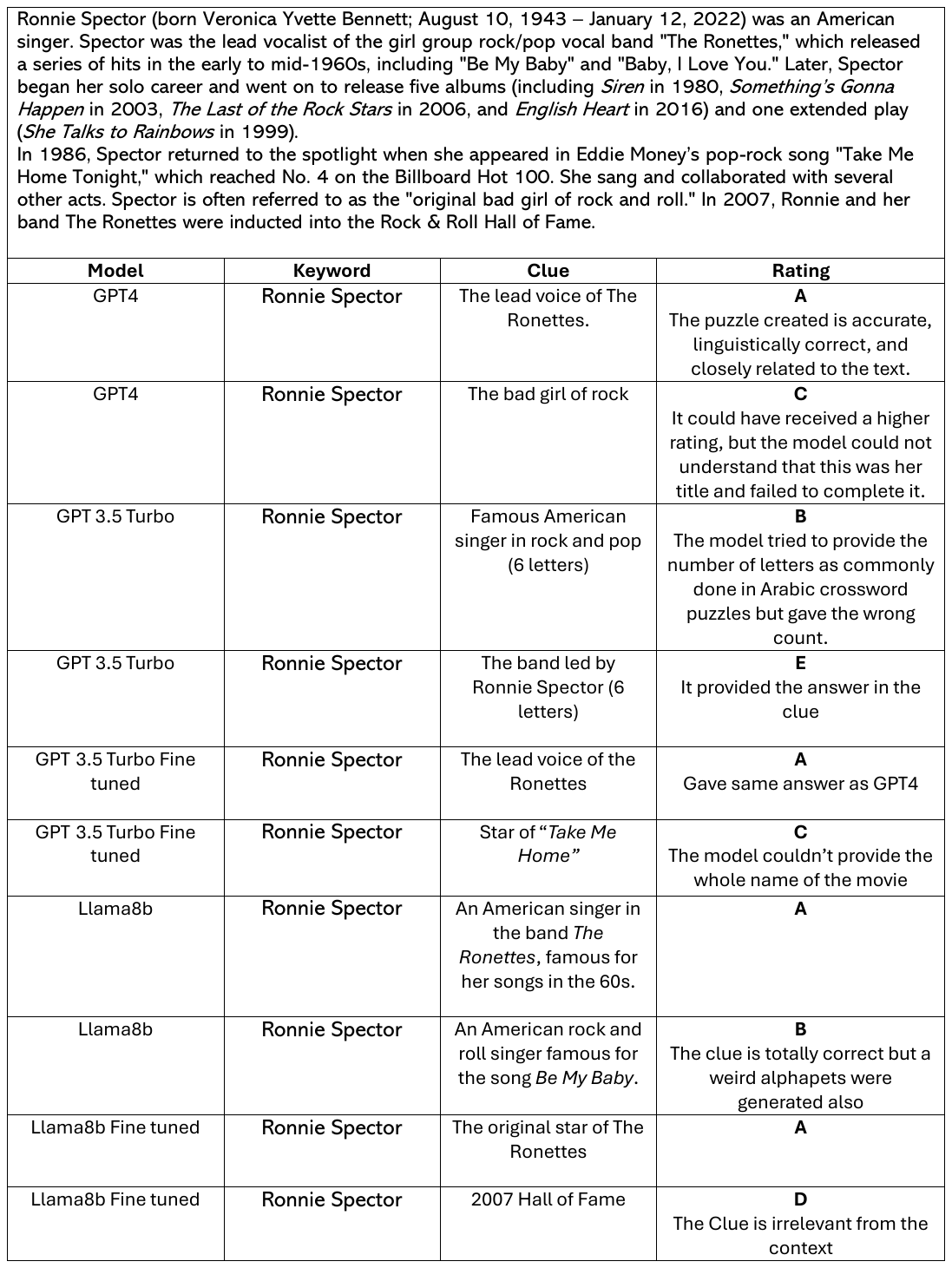}
   
    \caption{Example of Ronnie Spector article translated in English} \label{fig:Example1_EN}
\end{figure*}

\begin{figure*}
    \centering 
   \includegraphics[width=1\textwidth]{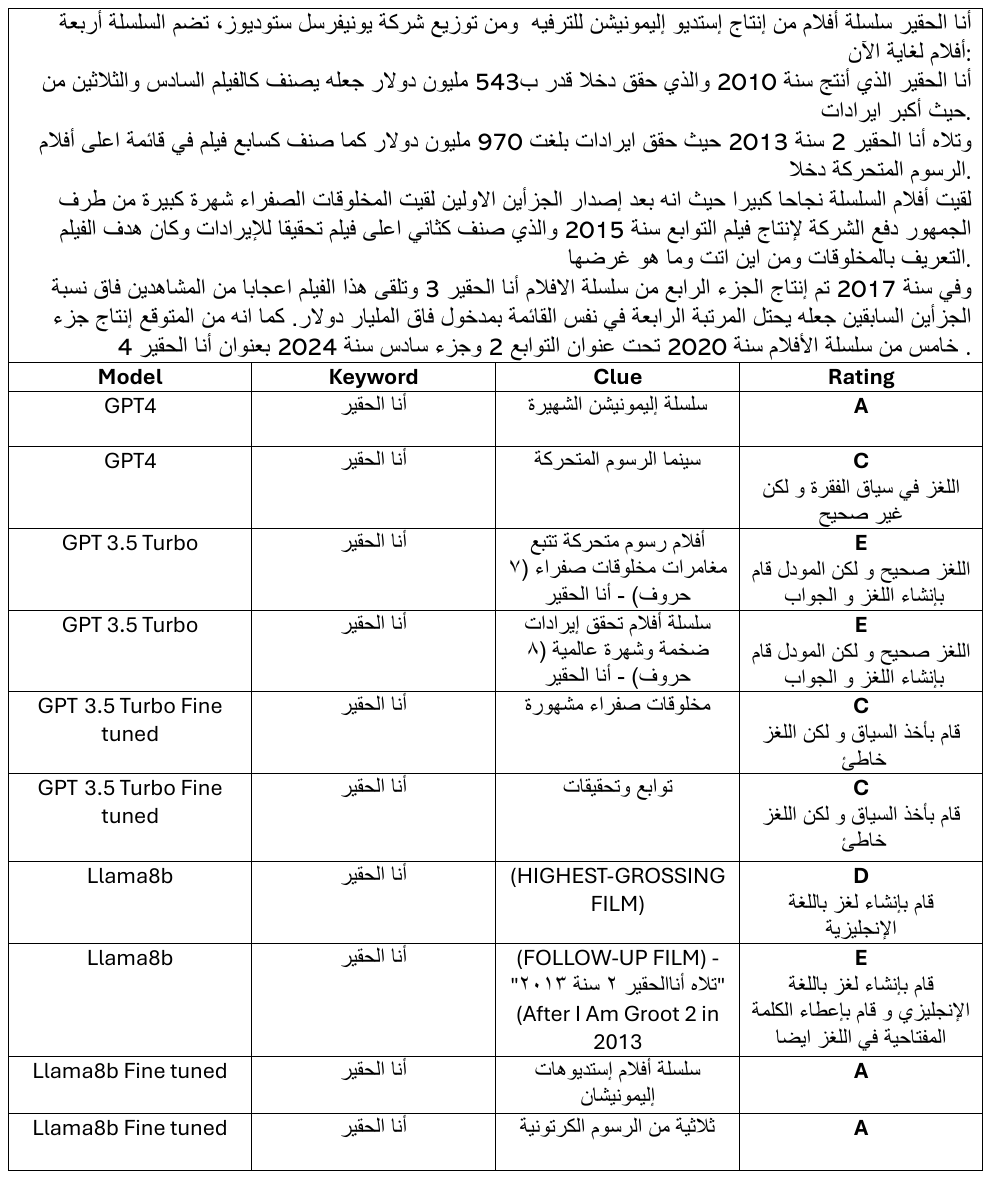}
    
    \caption{Example of Despicable Me article}\label{fig:Example2_AR}
\end{figure*}

\begin{figure*}
    \centering 
   \includegraphics[width=1\textwidth]{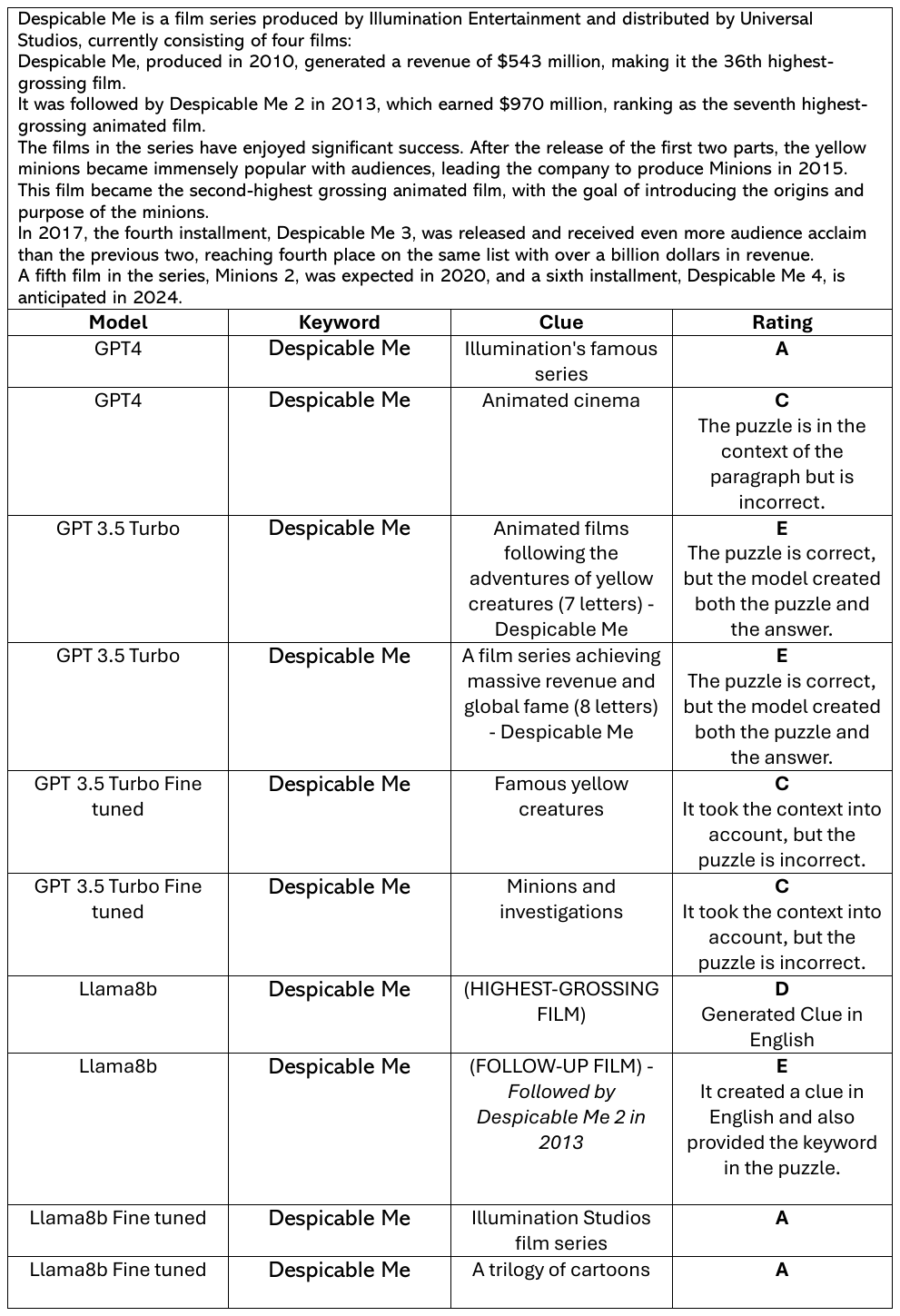}
    
    \caption{Example of Despicable Me article translated in English}\label{fig:Example2_EN}
\end{figure*}

\begin{figure*}
    \centering 
   \includegraphics[width=1\textwidth]{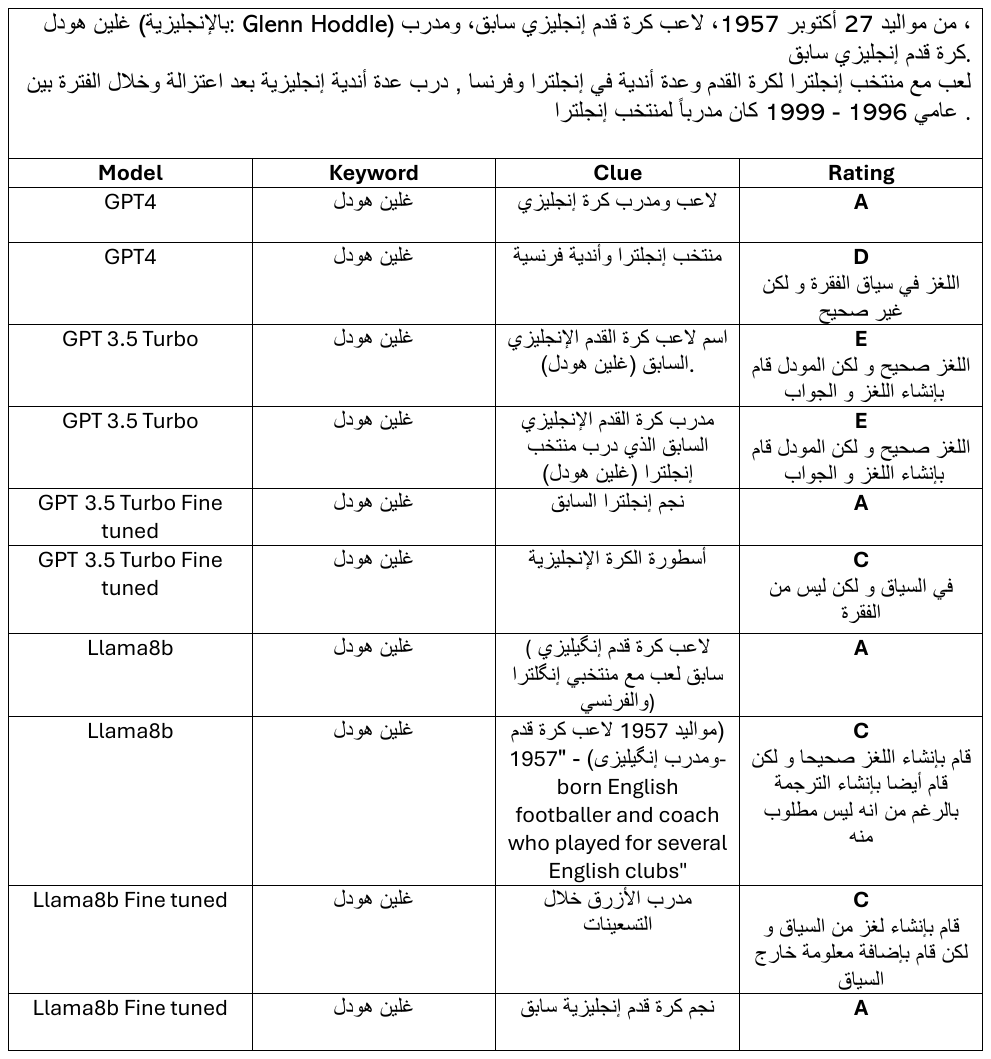}
    
    \caption{Example of glenn hoddle article}\label{fig:Example3_AR}
\end{figure*}

\begin{figure*}
    \centering 
   \includegraphics[width=1\textwidth]{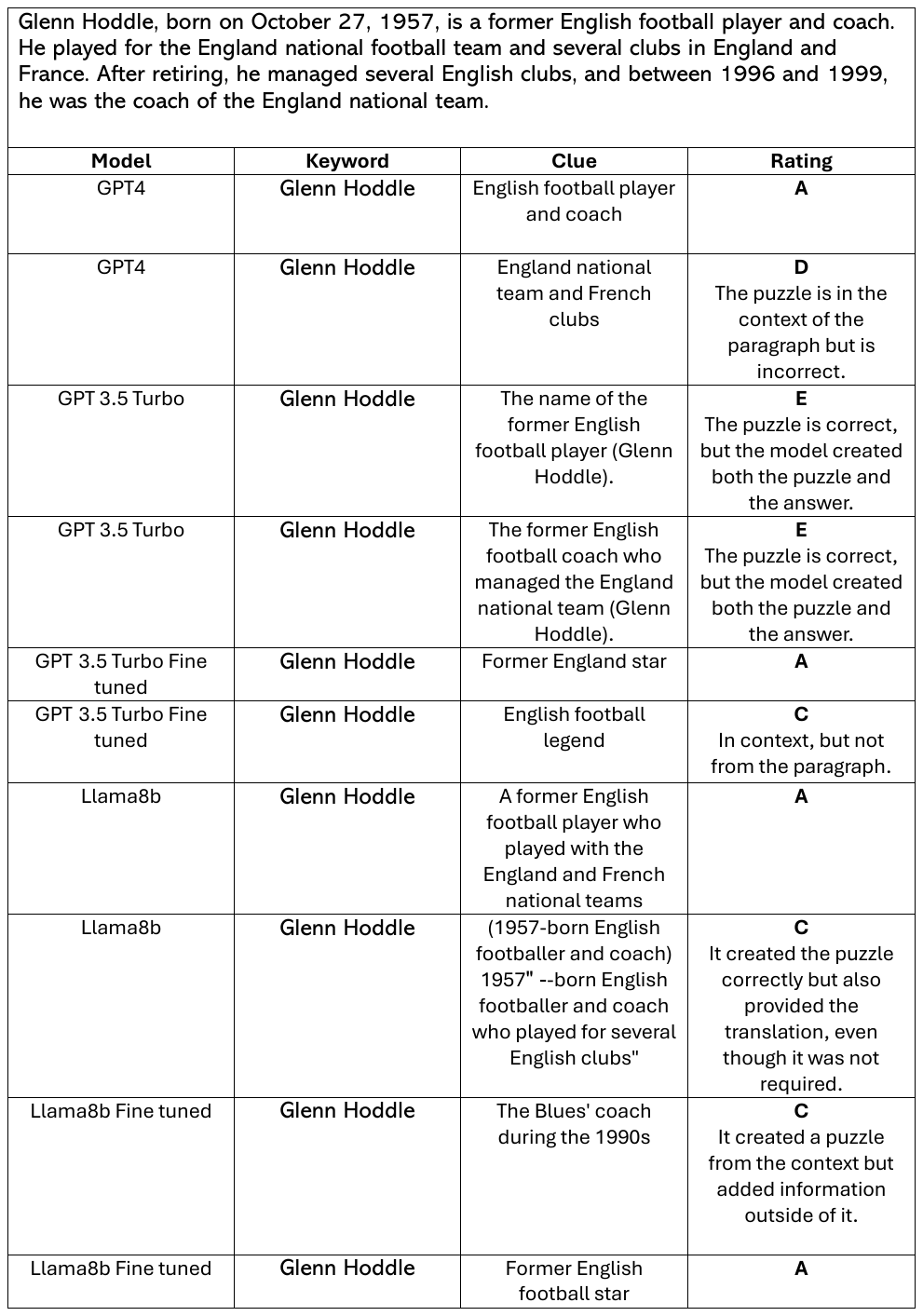}
    
    \caption{Example of glenn hoddle article translated in English}\label{fig:Example3_EN}

\end{figure*}

In this section, we present a selection of evaluated examples to illustrate the output generated by each model employed for clue generation, specifically: \texttt{GPT-4}, \texttt{GPT-3.5-Turbo}, \texttt{GPT-3.5-Turbo Fine-Tuned}, \texttt{Llama3-8B-Instruct}, and \texttt{Llama3-8B-Instruct Fine-Tuned}. The examples were carefully chosen to cover a range of ratings (A, B, C, D, and E) determined by expert evaluators in Arabic language proficiency. The examples were provided in \ref{fig:Example1_AR} , \ref{fig:Example2_AR} and \ref{fig:Example3_AR}, followed by their translation in \ref{fig:Example1_EN} , \ref{fig:Example2_EN} and \ref{fig:Example3_EN}

\newpage
\section{Human Observations and Analysis of LLMs on Clue Generation} 
\label{appendix_2}
Each model demonstrates specific failure patterns in generating clues from the given text. The \texttt{GPT-4} model, for instance, often struggles when keywords appear in brackets within the text. This limitation extends to instances where place names, movie titles, or other notable phrases are enclosed in brackets or quotations, as seen in \ref{fig:Example1_AR}. Such failure patterns include misinterpretations of context, along with occasional hallucinations.

The \texttt{GPT-3.5-Turbo} model displays different challenges. For example, it frequently attempts to offer letter counts as hints, a typical feature of Arabic crossword puzzles, but consistently fails to provide accurate counts. In Arabic crosswords, clues are conventionally labeled as "horizontal" ( afqi) or "vertical" ( r'asī), followed by the respective clues. Many D-labeled clues generated by this model simply state "horizontal:" (afqī:) without following with a proper clue. When the "keyword" is contained in brackets, as in \ref{fig:Example3_AR}, the model sometimes generates clues that inadvertently include the answer.
Similarly, \texttt{Llama-8b} encounters difficulty when Latin characters are present within the text, an issue clearly illustrated in Examples \ref{fig:Example1_AR}, \ref{fig:Example2_AR}, and \ref{fig:Example3_AR}.\\

While the base models \texttt{GPT-4}, \texttt{GPT-3.5-Turbo}, and \texttt{Llama-8b} exhibit specific failure patterns, the fine-tuned versions \texttt{GPT-3.5-Turbo Fine Tuned} and \texttt{Llama-8b-Fine Tuned} show a different issue. Although these fine-tuned models generate clues using the information within the text, they also tend to incorporate additional details from their broader knowledge base. This reliance on external knowledge, as observed in Example 3 (\ref{fig:Example3_EN}), can lead to inaccuracies or unintended context, thereby complicating the model’s ability to remain strictly within the provided text.
\end{document}